\documentclass[runningheads]{llncs}
\usepackage{amsmath}
\usepackage{multirow}
\usepackage{graphicx}
\usepackage{float}
\usepackage{booktabs}
\newcommand{\head}[1]{{%
  \normalfont\bfseries
  \begin{tabular}{@{}c@{}}
  #1
  \end{tabular}%
}}

\begin{document}
\title{Deceptive Kernel Function on Observations of Discrete POMDP}
%
%
\author{Zhili Zhang\inst{1}\orcidID{0000-0002-0870-3304} \and
Quanyan Zhu\inst{1}\orcidID{0000-0002-0008-2953} }
\authorrunning{Z. Zhang and Q. Zhu.}
%
\institute{New York University, New York City, NY 10003, USA \\
\email{\{zz2382,qz494\}@nyu.edu}}
\maketitle              
\begin{abstract}
This paper studies the deception applied on agent in a partially observable Markov decision process. We introduce deceptive kernel function (the kernel) applied to agent's observations in a discrete POMDP. Based on value iteration, value function approximation and POMCP three characteristic algorithms used by agent, we analyze its belief being misled by falsified observations as the kernel's outputs and anticipate its probable threat on agent's reward and potentially other performance. We validate our expectation and explore more detrimental effects of the deception by experimenting on two POMDP problems. The result shows that the kernel applied on agent's observation can affect its belief and substantially lower its resulting rewards; meantime certain implementation of the kernel could induce other abnormal behaviors by the agent.

\end{abstract}
%
%
\section{Introduction}
Deception as a concept has its meaning and utility wide across different disciplines and subjects~\cite{pawlick2019game}.  Within the field of cybersecurity, deception as a general strategy has been discussed frequently in the game-theoretic framework from the defender perspective. Carroll and Grosu~\cite{carroll2011game} have investigated the `honeypot' strategy as a deceptive approach taken by defenders and modeled the defender and attacker's behaviors after deployment of deception as a signaling game. Pawlick et al.~\cite{pawlick2018modeling} have proposed a model of deception based on cheap-talk signaling game with a probabilistic deception detector. Ding et al.~\cite{ding2020defensive} have given a dynamic game model for the deployment of deception against reactive jamming attacks. As stated by Pawlick et al. in the taxonomy and survey of defensive deception~\cite{pawlick2019game}, game theory provides an theoretical base for modeling of cybersecurity scenarios as it precisely describe the competitive and adversarial relationship between defender and attacker. 

Partially observable Markov decision process (POMDP) as a general framework of sequential decision process is also widely used in cybersecurity researches of defensive and attacking approaches. From defender's perspective, Miehling et al. in~\cite{miehling2018pomdp} have adopted POMDP to optimally interfere the attacker's progression; Hu et al. in~\cite{hu2017online} have also modeled the defense problem with POMDP and proposed online algorithm for adaptive defense measure. Yu and Brooks~\cite{yu2017stochastic} have applied POMDP and its approximation solution as a stochastic tool for network intrusion detection. On the other hand, Sarraute et al.~\cite{sarraute2012pomdps} in an attacker's view have introduced POMDP in attack planning issue of penetration testing. In either way, POMDP being a model concerning partial observability and probabilistic transition is a handful tool for cybersecurity researches and more researches related to decision-making.

However, as an ``overlapping'' field of deception strategy and POMDP's application areas, few research in cybersecurity has investigated the deceptive approaches in POMDP, or other stochastic sequential decision-making models. Hor{\'a}k et al. in~\cite{horak2017manipulating} have adopted the model of one-sided partially observable stochastic game (POSG) and discussed the proactive deceptive actions by the defender as an approach to manipulate the adversary's (the attacker) belief. In~\cite{ahmadi2018partially}, Ahmadi et al. have also formalized the deception game as one-sided POSG for further implementation of its solution. While both authors aims to put deception issue in a stochastic decision-making setting with partial observability, they have also stressed that the difficulty of this combination resides in the intractability of solving a general POSG, where the imperfect information of both parties presents substantial theoretical and computational challenges. On the other hand, POMDP modeling the decision process of a single agent is limited in describing adversarial scenario, which is suggested as a nature of deception in general. These factors limit the relative research in cybersecurity area, and further our understanding in general over an agent whose decisions follow POMDP model is meantime deceived.

Out of the box of cybersecurity, relative researches across deception and stochastic decision-making process are just emerging in optimal control and reinforcement learning areas. Ornik and Topcu in~\cite{ornik2018deception} have introduced a mathematical framework for deception in optimal control based on Markov decision process (MDP). However, the framework has a prerequisite of its deception objective being reward-based. Wu et al.~\cite{wu2019reward} have studied the agent's behavior within an MDP environment subject to a reward-based adversarial deception, which in the meantime leverages the cognitive bias of the human. More related researches have emerged recently in reinforcement learning (RL) area regarding the security and stability of RL process. Huang and Zhu~\cite{huang2019deceptive} have investigated the falsified cost signal in RL with Q-learning algorithm by analyzing its impact on algorithm's convergence and the mechanism of misleading effect by the cost falsification. Russo and Proutiere in~\cite{russo2019optimal} have introduced an optimal attack targeting a given policy of the RL agent by modeling this problem as an MDP and solve it with Deep RL techniques. Zhang et al. and Rakhsha et al. in their respective researches~\cite{zhang2020adaptive} and~\cite{rakhsha2020policy} have proposed a training-time poisoning attack targeting the environment (reward and dynamics of MDP) and specifically reward. Both researches aim to achieve policy manipulation/teaching through manipulation of original MDP to a `poisonous' MDP. While these researches all demonstrate novel and inspirational work from their respective angle, they are meantime subject to several common limitations. Attacks modeled in researches~\cite{ornik2018deception},~\cite{zhang2020adaptive} and~\cite{rakhsha2020policy} are subject to certain ``objectives'' of deception/attack; research in~\cite{wu2019reward} considers human cognitive bias as a reason for agent's potential suboptimal choices; models in~\cite{wu2019reward},~\cite{russo2019optimal},~\cite{zhang2020adaptive} and~\cite{rakhsha2020policy} are computed with underlying MDP regardless of agent's partial observability as a potential setting; and some of the tactics introduced above are more attacking rather than deceptive.

In POMDP, the agent naturally maintains a belief over environment's state as a brief and adequate substitution of entire `history', because of the absence of state information. As is quoted by Pawlick et al. in~\cite{pawlick2019game}, a definition of deception is given as ``to intentionally cause another person to acquire or continue to have a false belief, or to be prevented from acquiring or cease to have a true belief''. The coincidence of `belief' in both statements motivates our questions: ``how do we deceive an agent's belief in POMDP without a macro objective'' and ``what is the impact of this deception on agent's decision making in a POMDP''. Motivated by the achievements and limitations from aforementioned work, we hope to introduce a deceptive approach regarding a RL agent with partial observability. Specifically, regardless of the adversarial nature of deception and an objective-based attacking model, we apply a simple modification to agent's observation through the implementation of deceptive kernel function. We aim to evaluate its influence on agent's belief, and as a result its decisions and returns. Meantime, we also investigate how `successful' the deceptive approach is through the reflections of several indices. The main contributions of our paper can be summarized as follows:
\begin{itemize}
\item	We formalize the deceptive kernel function in POMDP and introduce three simple implementations.
\item	We investigate agent's belief and decisions under the impact of deception and provide theoretical reasoning of deceptive effect based on three different algorithms adopted by the agent. 
\item	We carry out experiments on two POMDP problems. By comparing statistics from baseline and experiments under deception, we evaluate the effectiveness of our approach. 
\end{itemize}

\section{Preliminaries and Problem Formulation}
\subsection{POMDP}
An MDP is a discrete-time stochastic control process. It provides a model of interactions between an environment and an agent, whose decision making satisfies the Markov Property, the Memoryless property of a stochastic process such that given a present state (time $t$ and state $k_t$), the past ($k_{t-i}$) and the future ($k_{t+i}$) are independent. An MDP can be formalized as a 5-tuple $\{S, A, T, R, \gamma\}$. $S$ is the set of states. $A$ is the set of actions. $T$ ($T(s, a, s')$) is the transition model that specifies the probability from state $s$ to state $s'$ taking action $a$. $R$ ($R(s, a, s')$) is the reward model that provides the stepwise reward provided with the state transition and action taken. 

A POMDP is a generalization of MDP such that its underlying dynamics is an MDP, but the transition model and the actual state information is hidden from the agent, where the partial observability stands for. This paper studies the discrete and finite POMDP, whose state space is a discrete and finite set. It can be formalized as a 7-tuple: $\{S, A, T, R, \Omega, O, \gamma\}$:
\begin{itemize}
	\item $S$ is a finite set of states.
	\item $A$ is a finite set of actions.
	\item $T$ is the transition model of the underlying MDP.
	\item $R$ is the reward model of the underlying MDP.
	\item $\Omega$ is the set of observations.
	\item $O: O(o | s', a)$ is the condtional observation model that specifies the probability of observation $o$ given state $s'$ and action $a$.
	\item $\gamma \in [0,1]$ is the discount factor portrays the stepwise reward loss.
\end{itemize}

Because of the absence of state information, the agent needs to maintain a belief $b$ of its state, a probability distribution over the state space.  The belief updating model along with the introduction of belief is the transition probability function of belief that follows the Bayes' rule. It can be formalized as below ($\eta$ is a normalizing constant):
\begin{equation}
b'(s') = \eta O(o |s', a) \sum_{s\in S}{T(s'|s, a)b(s),\ \eta = \frac{1}{Pr(o | b,a )}} 
\end{equation}
$$ Pr(o |b,a ) = \sum_{s'\in S} O(o |s', a) \sum_{s\in S}{T(s'|s, a)b(s)} $$

The introduction of belief and belief update allows us to transform a POMDP into a belief MDP, whose states are the belief states of the agent and the transition model is the combination of original transition, observation and belief update models. The belief, as its literal meaning of agent's knowledge and conception of its state, is intrinsically the target of deception, which to many extents stands for the approach and methodology that has one major objective of affecting and manipulating the target's belief. It will underly the major analysis of our proposed deceptive approach in this paper.
\subsection{Observation Model and Deceptive Kernel Function Specification}
Deceptive kernel function (abbr. kernel) is an approach we introduce that aims to deceive its target, in this case the agent in a POMDP, through manipulation of observation. The kernel stands for kernel function, that, with the input of the original observation $o$, outputs a new observation $o'$.
$$ Kernel: \Omega \rightarrow \Omega,\ o' = Kernel(o) $$
A straightforward yet inaccurate analogy of the kernel is sunglasses, which transform the input view to a filtered darker one. The kernel in our conception and experiment can do more, such as reversing the input observation, or masking the original and sampling an irrelevant one.
\subsubsection{Kernel Target} \hfill\\

The deceptive kernel targets the observation received by the agent. Generic observations in many real-life POMDP-like problems take various forms. The \textit{Tiger} Problem, a simple POMDP introduced by Kaelbling et al. in~\cite{kaelbling1998planning} and experimented in this paper provides a binary observation; a sound by the tiger (a probabilistic indicator of the presence of tiger) is heard from either door by the agent. In another experiment carried out in this paper with \textit{RockSample} Problem, a POMDP abstracted from rover exploration of robots~\cite{smith2012heuristic}, the observation set $\Omega$ is comprised of informative observations and uninformative ones. The former subset of observations is divided based on the observed target such that each target has a corresponding and unique binary observation space of being positive and negative. The latter, being ``uninformative'' on the other hand, cannot trigger a belief-update when it is perceived, and it maintains the irrelevance from agent's decision or state transition. 

In this paper, we focus on informative observations and ignore the uninformative ones from being a target of our kernel because they cannot contribute to the influence imposing on agent's belief. Decision process in real life allows for not only binary or finite discrete results of observation, but infinite and continuous ones.  In order to maintain the simplicity of computation, we only consider POMDP with discrete finite set of observations in our analysis and experiments. 

\subsubsection{Kernel Specification} \hfill\\

In many POMDP model abstracted and generated from real-life problems, we usually consider noise factor in the environment. The noise factor originates from the fact that agent's sensors, being human or machine sensations, are inherently noisy. It contributes to the innate deception that the natural environment exerts on us. Therefore, we apply a probabilistic observation model which can be formalized as below:

\paragraph{Define} $p_T \in [0,1]$ be the correct rate of our sensor outputting correct observation regarding a specific observed object; $p_F = 1-p_T$ be the noisy index.  Assume there is one single true observation $o_t$ and a set of possible fake possible observations $O_F$. The probabilistic model can be formalized as:
$$P(o | o = o_t) = p_T;\ P(o | o \in O_F) = p_F$$
\noindent Based on this model, we design and hereby introduce three kinds deceptive kernels, \textit{Prob}, \textit{Rand} and \textit{Oppo}, with respective deception mechanisms which can be formulized as below:

\paragraph{Prob} kernel: Define $p_k$ as the deceptive rate. The \textit{Prob} kernel serves as a filter that allows fewer correct observation at the ratio of $p_k$ provided with original noisy observation model.
\begin{equation}
P_{Prob}(o | o = o_t) = p_T \cdot p_k,\ P_{Prob}(O | o\in O_F) = 1 - p_T \cdot p_k 
\end{equation}

\paragraph{Rand} kernel: \textit{Rand} kernel outputs a uniformly random sample from the observation space: $\Omega = \{o_t\} \cup O_F$.
\begin{equation}
P_{Rand}(o | o = o_t) = \frac{1}{|\Omega|},\ P(o | o \in O_F) = 1 - \frac{1}{|\Omega |}
\end{equation}

\paragraph{Oppo} kernel: \textit{Oppo} kernel deterministically outputs the false observation.
\begin{equation}
P_{Oppo}(o | o = o_t) = 0,\ P(o | o \in O_F) = 1
\end{equation}

\paragraph{Remark 1.} 
From computational perspective, we can treat the \textit{Rand} kernel and \textit{Oppo} kernel as special cases of \textit{Prob} kernel with respectively $P_{Prob} = \frac{1}{|\Omega|}$ and $P_{Prob} = 0$. In other words, all three kernels are probabilistic kernels with different `degrees'. The underlying reason of this categorization is the difference in the implementations of kernels, primarily the programming implementations. The \textit{Rand} kernel was implemented with an observation sampler which ignores the original input; the other two was realized through masking strategy with their respective filtering rate $P_{Prob}$. The difference also imply the actual deception strategy in real-life implementations. The \textit{Rand} kernel can target the agent by blinding its sensor, while the others are more likely targeting the environment factors such as introducing more artificial noise, transforming signals from agent's observed object or `honeypot' strategy~\cite{carroll2011game}.

\paragraph{Remark 2.} 
Regarding the trivial difference between two POMDP problems studied in this paper, the \textit{Tiger} problem adopts a simple probabilistic observation model as mentioned in~\cite{kaelbling1998planning} with $p = 0.85$. The \textit{RockSample} problem however adopts a slight variance of the model such that the correct rate $p$ is not a constant, but a variable satisfying: 
\begin{equation}
p = \frac{1}{2} \cdot (1 + 2^{-\frac{d}{D} }  )
\end{equation}
$d$ is the distance from agent to the target rock; $D$ is the constant marking the distance of $0.75$ correct rate and has value $20.0$ in our experiment.

This original setting aims to consider the actual performance attributes of detecting sensors whose detecting accuracy decreases when they target farther objects. From our experimental purpose to distinguish between \textit{Prob} kernel and the others with regard to deception effect, we adopt the deceptive rate $p_k$ in our experiments to satisfy that the aggregate deceptive rate of natural environmental noise and the \textit{Prob} kernel is no worse than the \textit{Rand} kernel in both POMDPs, especially for the worst case in \textit{RockSample} problem. Briefly, we guarantee in our experiment that $P_{prob} > P_{Rand} > P_{Oppo} = 0$ with regard to aggregate correct rates $P_{prob}, P_{Rand}, P_{Oppo}$ of outputting true observation respectively for three kernels. 

\subsubsection{Cost for Deception} \hfill\\

In the experiment with \textit{RockSample} problem, we further introduce the cost of deception as a considered factor. However, different from the cases in~\cite{carroll2011game} where the authors models the problem as a signaling game with two parties – the attacker and the defender, there is no adversary to the agent in POMDP problem. The essential target of our deceptive kernel is the environment regardless of the implementation. The kernel aims to ``modify'' the environment to be noiser and more deceptive. And this modification is less likely to be free of charge in real life. To implement this idea, we treat the cost of deception as a tiny reward to the agent. Specifically, we introduce a stepwise reward $r_d$ associated with each application of deceptive kernel and combine the reward with original rewarding model of the problem. 

\paragraph{Remark 3.} 
The probabilistic observation model allows the cases of false observation received by the agent as a result of environmental noise. Meantime, the probabilistic deceptive kernels also allows correct observation as long as the aggregate deceptive rate is greater than 0. The combination of two models raises a special case where natural observation acquired by the agent is false while the observation processed by deceptive kernel is true. This also alludes a real case that the agent can be wrong with what it observes; a random guess can be correct on the contrary. Therefore, we define two Boolean flags \textit{isFalse} and \textit{isDeceived} regarding the observations $o_d = Kernel(o)$ which is processed with deceptive kernel, provided with the original observation $o$ (which would have been perceived by the agent without kernel).

\paragraph{Define} the single true observation is $o_t$, (the set of false observations is $O_F$), the original observation is $o$:
\begin{align*}
isFalse(o_d) &= (o_d \neq o_t) \\
isDeceived(o_d) &= (o_d \neq o)
\end{align*}

\section{Analysis of POMDPs Applied with Deceptive Kernel}

In this section, we investigate the application of deceptive kernel on two simple POMDP problems through the theoretical approach. Particularly, we borrow a common case of observation model appeared in both problems experimented in this paper and evaluate the discrepancy of the resulting beliefs after receiving the correct and incorrect observations. Provided that agent's belief is directly affected by the deceptive observations, we assume the agent can further receive collateral damage, potentially in its actions, rewards and other aspects of decision-making process. One example of damage could be a significantly prolonged duration of game-play by the agent, who wanders around inside the maze as being deceived. Then, based on the algorithms used by the agent in either problem, we evaluate these potential detrimental effects to the player's decision-making that can be indirectly attributed to the  deception.

\subsection{Belief Affected by Deception}
We re-visit the belief-update rule in POMDP in formula (1).
\begin{equation*}
b'(s') = \eta O(o |s', a) \sum_{s\in S}{T(s'|s, a)b(s),\ \eta = \frac{1}{Pr(o | b,a )}} 
\end{equation*}
The new belief of the agent is jointly determined by the following factors: agent's prior belief, transition model and observation probabilistic model of the POMDP. The formula considers the complex possibilities of belief updating in theory. Now we simplify the belief updating model in multiple approaches. 

We first limit the represented target of state $s$ to a single object with binary state instead of the entire environment. In the case of \textit{Tiger} problem, the state of entire environment is binary. In the more complex case of \textit{RockSample}, the state of environment can be split into two parts, the agent's position and the rocks' states. The former is independent from agent's observation model and the belief updating model, primarily because the agent's position is purely controlled by the agent's deterministic moving actions, i.e., agent is not allowed to make illegal moving action that leads to a forbidden position; all the legal moving actions will deterministically get the agent to the target position. The latter can be further disassembled into eight binary states corresponding to eight fixed and known rock positions in the map. States in real life are more likely to be continuous and comprehensive instead of singular or discrete. However, treating the state $s$ as a singular binary descriptor reasonably simplifies the computation and allows us to carry out analysis on belief towards a single observed target. 

Further, we categorize the action set $A$ into two subsets for operating actions $A_{op}$ and observing actions $A_{ob}$. The former set contains actions that only transit the environment's state but do not generate an observation or induce a belief-update. The latter set contains actions that only generate an observation but do not transit the state. 
\paragraph{Define} $a_{op}$ is an operating action if, by taking $a_{op}$, the environment transits from state $s$ to $s'$, while the agent's belief remains as previous.
$$ b(s') = b(s) $$ 
\paragraph{Define} $a_{ob}$ is an observing action if, by taking $a_{ob}$ at time $k$, the environment maintains its previous state $s_k$, and the agent receives a binary observation $o_{ob}$
\begin{align*}
& For\ s_{k+1} = s_k,\ T(s_{k+1} | s, a_{ob}) = 1,\ O(o_{ob} | s_k, a_{ob}, s_{k+1}) >= 0; \\
& for\ s_{k+1} \neq s_k,\ T(s_{k+1} | s, a_{ob}) = 0,\ O(o_{ob} | s_k, a_{ob}, s_{k+1}) = 0
\end{align*}

\paragraph{Remark 4.}
The purpose of introducing categorization of actions is not to redefine all actions and steps in POMDP, but to distinguish the `belief-update' stage from the `operation' stage. A generic step in POMDP consists of action-taking, state-transiting, observing and belief-updating processes. An instance of explorative rover robot and possibly many real-life robots shows that the detecting actions taken by their sensor works independently from their dynamic operations, even though both functionalities share the same control flow. By splitting at the middle of transiting and observing, we aim to create a stage of observation and belief state transition independent from the environment's state transition. We want to focus on agent's belief update resulted purely from receiving deceptive observation, and evaluate the difference between the deceived belief and the true belief.

Now we define $s$ a binary state with value either $0$ or $1$. After the agent performs an observing action, it can either receive the true observation $o_T$ at probability $p_T$, or the false observation $o_F$ at probability $p_F$. $p_F = 1 - p_T$.

\paragraph{Define} the agent's current belief towards binary state $s$ is $b(s) = (p_0, p_1)$. Without loss of generality, we assume the true state of $s$ is $s = 0$. Hence, $o_T = o_0$, $o_F = o_1$. 

\noindent Having received the true observation, the observation probabilities are:
$$ O(o_T | s = 0) = p_T,\ O(o_T | s = 1) = p_F $$
\noindent The true belief $b'(s)$ after belief updating is:
\begin{align*}
b'(s = 0) &= \frac{ O(o_T | s = 0)\cdot b(s=0)}{ O(o_T | s = 0)\cdot b(s=0) + O(o_T | s = 1)\cdot b(s=1)} \\
b'(s = 1) &= \frac{ O(o_T | s = 1)\cdot b(s=1)}{ O(o_T | s = 0)\cdot b(s=0) + O(o_T | s = 1)\cdot b(s=1)}
\end{align*}
\noindent On the other hand, the observation probabilities having received the false observation are:
$$ O(o_F | s = 0) = p_F, O(o_F | s = 1) = p_T $$
\noindent The deceived belief $b''(s)$ after belief updating is:
\begin{align*}
b''(s = 0) &= \frac{ O(o_F | s = 0) \cdot b(s=0)}{ O(o_F | s = 0) \cdot b(s=0) + O(o_F | s = 1) \cdot b(s=1)}\\
b''(s = 1) &= \frac{ O(o_F | s = 1) \cdot b(s=1)}{ O(o_F | s = 0) \cdot b(s=0) + O(o_F | s = 1) \cdot b(s=1)}
\end{align*}
\noindent We evaluate the difference of agent's belief towards $s=0$ between both cases:
\begin{align*}
\frac{b'(s = 0)}{b''(s = 0)} &= {\frac{ p_T \cdot p_0}{ p_T \cdot p_0 + p_F \cdot p_1}} \cdot \left(\frac{ p_F \cdot p_0}{ p_F \cdot p_0 + p_T \cdot p_1}\right)^{-1} \\
	&= \frac{p_T (p_F \cdot p_0 + p_T \cdot (1 - p_0))}{p_F (p_T \cdot p_0 + p_F \cdot (1 - p_0))} \\
	&= \frac{p_T^2 - (p_T - p_F)p_T p_0}{p_F^2 + (p_T - p_F)p_F p_0}
\end{align*}
As the $p_T$, $p_F$ being constants with assumption that $p_T>p_F$ and $p_T + p_F = 1$, we can easily see the above ratio monotonously decrease as $p_0$ ranges from $0$ to $1$. Then we get:
\begin{equation}
\frac{b'(s = 0)}{b''(s = 0)} \in [1, (\frac{p_T}{p_F})^2]
\end{equation}
The result shows that the difference between a true belief and a deceived belief can be as large as $(\frac{p_T}{p_F})^2$ times. In the experiment of \textit{Tiger} problem and \textit{RockSample} problem, we adopt $p_T = 0.85$. The upper bound of discrepancy being $32.11$ is outstanding.

\subsection{Decision Affected by Deception}
Actions taken by the agent are primarily decided by the agent's policy, which is generated by its adopted solutions. Solutions to POMDPs vary significantly across their settings, strategies and even specific stepwise approaches. However, since the state information is not available, the agent must rely on what is observable and consider the history of its past actions and observations in deciding its next move. Belief state as the carrier of history plays a crucial role in deciding which action is optimal at current step. A deceived agent misled by false observation are more likely suffering from its unwise actions because the agent makes its `supposed' optimal choice on top of the incorrect basis, its belief. As the specific action-selection mechanism varies across different POMDP solutions or planning methods, it is difficult for us to enumerate all solutions to rationalize the detrimental effect resulted from deceptive observations. In the following section, we investigate three solutions to POMDP, the Value Iteration~\cite{smallwood1973optimal}, the Linear Value Function Approximation~\cite{ross2008online,hauskrecht2000value} and the POMCP~\cite{silver2010monte}. These algorithms were proposed by previous researches in chronological order and are correspondingly an optimal value function algorithm, an offline approximate algorithm and an online planning algorithm. We provide a simple and illustrative reasoning of misleading and detrimental effect from deceived incorrect belief respectively for these mentioned algorithms.

\subsubsection{Value Iteration and Linear Value Function Approximation} \hfill\\

POMDP can be solved optimally for a specified finite horizon $H$ by using the value iteration algorithm~\cite{smallwood1973optimal}. Further researches by Smallwood and Sondik~\cite{smallwood1973optimal} shows that optimal value function for a finite-horizon POMDP can be represented by hyperplanes, which are often called $\alpha$-vectors~\cite{ross2008online}. Each $\alpha$-vector is associated with some action and the vector defines the linear value function over belief space. By computing and finding the $\alpha$-vector mapping a specified belief to the maximum value, one can select the optimal action that corresponds to this vector. 

Value iteration algorithm computes $\alpha$-vectors by recursively applying the \textit{Bellman update} as the horizon increases. Although it provides an exact value computing approach for POMDP, the required space for $\alpha$-vectors grows exponentially as the algorithm iterates~\cite{ross2008online}. This brings about the introduction of approximate algorithms for value function. In the Linear Value Function approximation method, we treat $\alpha$-vectors as parameterized linear function over belief space and use a single-layer neural network to approximate the parameters of the function, essentially the $\alpha$-vectors. In both solutions, we aim to compute or approximate the $\alpha$-vectors because they can lead the agent to its optimal action whose corresponding $\alpha$-vector maps the provided belief to maximum value. 

\begin{figure}[ht]
\centering
\includegraphics[width=3.6in]{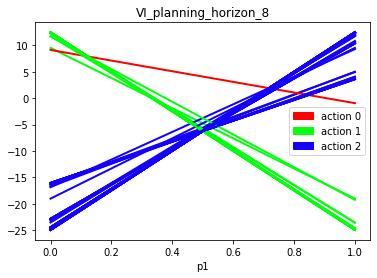}
\caption{$\alpha$-vectors applying value iteration at the horizon $H=8$ on \textit{Tiger} problem} \label{fig1}
\end{figure}

An illustration of resulting $\alpha$-vectors applying value iteration at the horizon $H=8$ on the Tiger problem is given by Patrick in his project as Figure \ref{fig1}~\cite{pomdpy}. We can visualize the $\alpha$-vectors and, moreover, the vector mapping a belief state to its maximum value and its counterpart optimal action. Meantime, we consider the preliminary analysis on the deceived agent's belief compared with its true belief. As the belief state of $s = 0$ traverses from the 0 to 1, its corresponding optimal action shifts among the action space. A deceived belief being diverged from its true value can be mapped to a wrong optimal action. The deceptive observation proves its utility in diverging agent's belief that possibly leads to a nonoptimal decision made by agent.

Furthermore, we can consider other effect from a special case of deceptive observation. The \textit{Rand} deceptive kernel mentioned in \textbf{Kernel Specification} section ``blinds'' the agent by feeding the agent randomly generated observation as counterfeit. Its application in \textit{Tiger} POMDP can lead the agent to a `trap' of repeatedly using one single action. The reason is that the agent is initialized with an impartial prior belief of the binary state with value $(p_0, p_1)=(0.5, 0.5)$. The Figure 1 shows that at initial state the agent is destined to \textit{listen} (action 0) because it needs observations to update its belief. A correctly diverged belief can guide the agent to open the correct door, either by taking action 1 or 2. However, with \textit{Rand} kernel applied, the agent could be repeatedly fed with contradictive observations, resulting in its belief to swing around the impartial state rather than diverge to either side. Therefore, the agent can be ``mentally'' trapped by disbelief in either side and repeatedly use \textit{listen}, meantime suffering the tiny cost affiliated to \textit{listen} until its belief eventually diverges.

To theoretically rationalize this effect of prolonging the duration by trapping the agent's belief with repeatedly contradictive observation, we assume a simple example where the original observation model gives correct observation at probability $p_T > 0.5$, and the agent's initial belief is $(p_0, p_1) = (0.5, 0.5)$. The agent will stop taking the \textit{listen} action and escape from trap until either $p_0$ or $p_1$ is larger than $p_T$. We assume the binary observation is either $T$ or $F$. 

Then, with one single observation, the agent cannot escape. The agent can escape after two observing actions if it receives either $[T, T]$ or $[F, F]$; it cannot escape otherwise. The agent cannot escape after 4 consecutive observations if primarily it does not escape after first two steps, and the last two observations it receives are either $[T, F]$ or $[F, T]$. We can simply calculate the probability of failure to escape after 2 steps and 4 steps as $P_2(fail)$ and $P_4(fail)$ respectively:
\begin{align*}
P_2(fail) &= 2p_T(1-p_T) \\
P_4(fail) &= [2p_T(1-p_T)]^2
\end{align*}
\noindent With original setting of $p_T = 0.85$, we can get:
$$	P_2(fail)= 0.255,\ P_4(fail) = 0.065 $$
\noindent With \textit{Rand} kernel applied and $p_T = 0.5$, we have:
$$ P'_2(fail) = 0.5,\ P'_4(fail) = 0.25 $$
\noindent The difference reflects that with \textit{Rand} kernel, the agent is more likely trapped by the misinformation from deceptive observations.

\subsubsection{Partially Observable Monte-Carlo Planning (POMCP)} \hfill\\

POMCP is the extension of Monte-Carlo tree search to partially observable environments~\cite{silver2010monte}. It consists of a UCT (Upper Confidence bound for Tree) search and an unweighted particle filter. A core contribution of the method is the introduction of search tree of histories, whose node $T(h)$ is defined by its corresponding history $h$ as $T(h) = <N(h), V(h), B(h)>$. $N(h)$ and $V(h)$ are the visit counter and the value of $h$ respectively; they are used by UCT algorithm to select the optimal action and are updated by simulator. $B(h)$ is the set of state particles used to approximate the belief state of $h$. At each step a real action is executed and a real observation is observed, particles inside $B(h)$ will be passed into simulator and filtered; particles whose simulated observation matches the real observation get to maintain in the set and are passed to the child history node. For detailed description of the algorithm, see~\cite{silver2010monte}; for details regarding implementation of the algorithm, see~\cite{pomdpy}. 

The reasoning behind the deceptive effect on agent's action selection is even more complicated compared to the previous case of value iteration, where the optimality is achieved by the action of which the corresponding $\alpha$-vector gives the maximum value for the given belief state. The UCT algorithm considers each actions' value and its corresponding visit count, with underlying reason that the algorithm aims to balance the exploration and the exploitation. The value of an action is approximated by the mean $q$-value of the action and is computed through Monte-Carlo simulations. We consider the pool of belief state particles $B(h)$ of history $h$ which is affected by deceptive observations throughout. Particles inside $B(h)$ are more likely prone to the deceived belief state rather than its true belief. This can be seen from the fact that particles are filtered by real observations which are potentially falsified. As these particles being sampled during Monte-Carlo Simulation and input to the simulator, the expected return is relatively lower if the rewarding model encourages true belief and penalizes false ones to some extent. 

In the example of \textit{RockSample} problem, the \textit{Sample} action is the only action that induces reward (except from entering the terminal state). It returns a positive reward only if agent's belief towards the targeting rock matches the actual state and that the rock must be good. On the other hand, it returns a penalty if the belief state does not match the actual state of the rock, regardless of what the actual state is. Combining with the previous analysis of stepwise belief, we see that deceptive kernels generating falsified observations are theoretically effective in this problem.

\section{Experiment with Two POMDP Problems}

In this section, we investigate two POMDP problems, the \textit{Tiger} Problem and the \textit{RockSample} Problem. The agent in the \textit{Tiger} Problem applies the Linear Function Approximation method~\cite{ross2008online,hauskrecht2000value} on the $\alpha$-vectors. In the latter problem, the agent adopts POMCP, a Monte-Carlo online planning algorithm introduced in previous section.

\subsection{Tiger Problem}

The Tiger problem is a single-player game with a simple setting. The agent is in front of two closed doors, one of which has a tiger behind. The underlying MDP has the state space $S$ with two states $\{TigerLeft, TigerRight\}$. The agent has three actions $\{Listen, OpenLeft, OpenRight\}$. By \textit{listening}, the agent will hear a sound either from left or right door as an indication of tiger behind. Yet, this observation is correct at probability $p_T = 0.85$. By \textit{opening} either door, the agent will subsequently enter the terminal state. The \textit{Listen} action has a tiny penalty $R(Listen) = -1$. Opening the door without tiger behind will result in a reward $R(safe) = 10$. Otherwise, the agent will receive a larger penalty $R(dangerous) = -20$. 

We applied three kinds of deceptive mask as defined in \textbf{Kernel Specification}. We assume for one \textit{Listen} step the true observation is $o_T$ and the false one is $o_F$; the agent receives $o$. Considering the environment's noise and kernel's deception, the aggregate probabilities of the agent observing correctly affected by each kernel are $P_{Prob}(o = o_T) = 0.6$, $P_{Rand}(O = o_T) = 0.5$, $P_{Oppo}(o = o_T) = 0$ respectively. 

\begin{table}[]
\centering
\caption{Statistics of Tiger Problem.}\label{tab1}
\begin{tabular}{c|c|c|c|c}
\toprule
                        & \head{Baseline}  & \head{Prob}      & \head{Rand}      & \head{Oppo}       \\
\midrule 
Undiscounted return     & $4.90 \pm 0.30$  & $-3.01 \pm 0.33$ & $-9.96 \pm 0.54$ & $-20.47 \pm 0.10$ \\ 
Discounted return       & $4.42 \pm 0.28$  & $-2.91 \pm 0.31$ & $-7.61 \pm 0.42$ & $-19.50 \pm 0.10$ \\ 
\# of Correct choices   & 895              & 623              & 488              & 0                 \\ 
\# of incorrect choices & 105              & 376              & 479              & 999               \\ 
Other cases             & 0                & 1                & 33               & 1                 \\ 
Average listen times    & 1.33             & 1.42             & 5.20             & 1.28              \\
\bottomrule
\end{tabular}
\end{table}

We carry out baseline experiment with original POMDP, POMDPs applied with \textit{Prob} kernel, \textit{Rand} kernel and \textit{Oppo} kernel. During each run for above POMDP models we investigate, we train the linear alpha vectors for 4500 epochs and run a validation epoch every 10th epochs while collecting data from it. We present the statistics of four sets of experiments in Table \ref{tab1}. It first compares the undiscounted and discounted returns (with discount factor $\gamma = 0.95$) from 1000 validation epochs of two runs for each model. We see the negative influence on agent's return by the application of deceptive kernel, and the influence get worse as the kernel becomes more deadly from \textit{Prob} to \textit{Oppo}. Also, in Table \ref{tab1} we show the agent's final decisions' distribution by the agent as for being correct or not. In the baseline model, the agent manages to open the safe door at approximately 90\% rate. This rate drops to 62\% and 50\% with the application of kernel. We can also notice particularly that the agent averagely listens for 5.2 times with the application of \textit{Rand} kernel, while this index has value reliably less than 2 for three other cases. This confirms our analysis in section 3.2 that the agent under this kernel are more probably trapped in a certain range of belief state, so that the agent has to repeatedly listen, suffering a longer pain. In extreme cases, we see the agent cannot decide the door within 50 steps and is forced out eventually. 

\begin{figure}[h]
\centering
\includegraphics[width=4.8in]{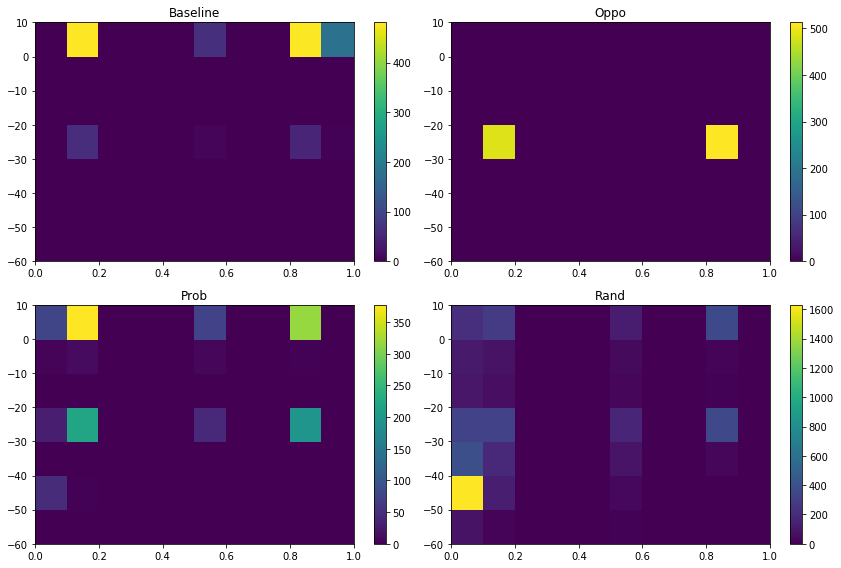}
\caption{2D histogram of agent's final reward over its belief state} \label{fig2}
\end{figure}

In Figure \ref{fig2}, we further plot the 2D histogram of agent's final reward with its belief state (the probability of tiger behind door 0). Particularly we enumerate not only agent's final belief, but its stepwise belief during decision process. The figure shows that in most epochs of baseline model, the agent listens once or twice and select the door. Although under the \textit{Prob} kernel the agent behaves similarly, it fails in more runs that lead to $-20$ penalty. From another perspective, we see that under \textit{Rand} kernel the distribution is more scattered along the Y-axis which represents agent's return. This, combined with the index of average listening steps, shows that agent must listen significantly more times to reach a decision while the aggregate tiny penalties from \textit{Listening} create this dispersive distribution of final return. The \textit{Oppo} kernel on the other extreme drives the agent strictly to listen one single time before selecting the definite wrong door. 

\begin{figure}[h]
\centering
\includegraphics[width=4.8in]{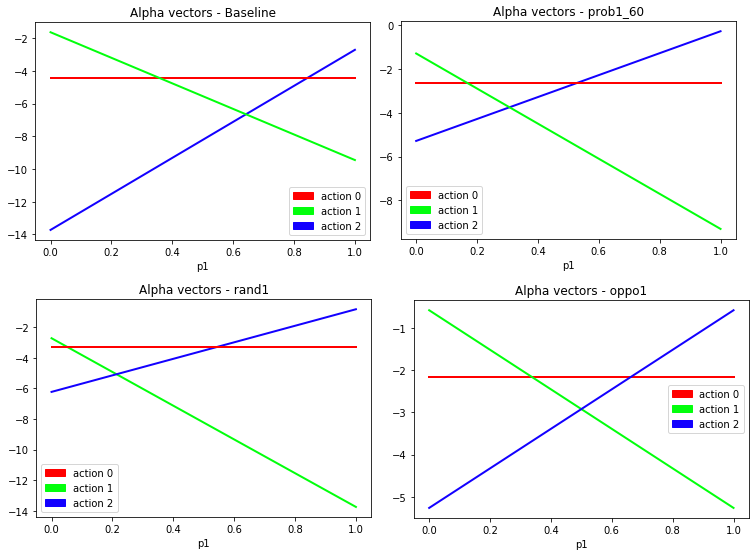}
\caption{Resulting $\alpha$-vectors in four experiments} \label{fig3}
\end{figure}

We also plot the four experiments' resulting $\alpha$-vectors in Figure \ref{fig3} to support the result.

\subsection{RockSample Problem}

\textit{RockSample} is a problem that models rover science exploration~\cite{smith2012heuristic}. The agent (rover) can traverse inside a map while sampling rocks at its immediate location. The agent's position and movement are deterministic and known to itself. The rocks' locations are also known by the agent, yet each rock has a binary state either being \textit{Good} or \textit{Bad}. The agent also maintains a binary belief over every rock and can only benefit from sampling a \textit{Good} rock that is also believed to be \textit{Good} by the agent. Otherwise the agent will be penalized by sampling either a bad rock or a believed-to-be-bad rock. 

\begin{figure}[h]
\centering
\includegraphics[width=2.5in]{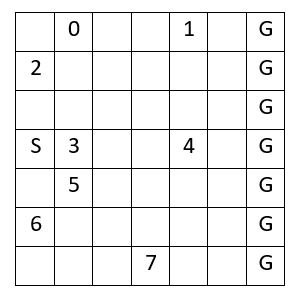}
\caption{Map of \textit{RockSample} Problem} \label{fig4}
\end{figure}

The map we adopt in experiment is given in Figure \ref{fig4}. ``S'' on the map marks the start position $(3,0)$; ``G'' marks the exit area where the agent enters the exit state in the coming step. Integer `i's ($i \in [0,7]$) marks the 8 rock positions. The action space $A$ consists of 13 actions $A = \{North, South, East, West, Sample, Check\-i\},\ i\in [0,1,…,7]$. The state space is the Cartesian product of locations and rock states $r = [r_0, r_1,\dots, r_7]$ where $r_i$ being $1$ or $0$ represents $i\-th$ rock is \textit{Good} or \textit{Bad} respectively. The observation corresponding to each \textit{Check-i} action is also binary, at the probability $p_T$ of being the correct observation. $p_T$ is correlated to the distance between the agent and the targeting rock, defined in formula (5). For the three kernels, the corresponding rates of outputting true observation are given in formula (2), (3) and (4) with $p_k = 0.8$ in \textit{Prob} kernel. 

\subsubsection{Experiment Results of RockSample} \hfill\\

For experiment with each kernel and baseline model, we execute the program for 500 epochs with randomly generated initial rock states, fixed lower and upper bound for belief state particles in each belief node and fixed amount of simulations for each Monte-Carlo tree search. We also carry out parallel comparison experiments with more simulations and larger particle sets $B(h)$ in each belief node $T(h)$. The result shows that the increase in these factors has very limited influence on the agent's return or decision-making. We provide in Table \ref{tab2} the statistics on the undiscounted and discounted return, average number of rocks sampled and checked, and average number of steps in agents' decision process for experiment with each setting.

\begin{table}[]
\caption{Statistics of RockSample Problem.}\label{tab2}
\centering
\begin{tabular}{c|c|c|c|c}
\toprule 
                      & \head{Baseline}  & \head{Prob}      & \head{Rand}      & \head{Oppo}      \\
\midrule
Undiscounted return   & $23.42 \pm 0.34$ & $21.34 \pm 0.30$ & $15.68 \pm 0.27$ & $10.04 \pm 0.03$ \\ 
Discounted return     & $15.26 \pm 0.19$ & $13.78 \pm 0.18$ & $8.93 \pm 0.17$  & $5.68 \pm 0.06$  \\ 
Average sampled rocks & 1.34             & 1.15             & 0.57             & 0.00             \\ 
Average checkings     & 4.37             & 4.50             & 5.74             & 5.57             \\ 
Average steps         & 14.21            & 13.88            & 14.65            & 12.86            \\
\bottomrule
\end{tabular}
\end{table}

The decrease in agent's both return indices for models carried with deceptive kernel is noticeable, which is also supported by the fewer numbers of \textit{Sample} actions made by the agent. From the index of agent's average steps, we see a similar effect of \textit{Rand} kernel that the agent tends to make more observing actions of \textit{Check} as the \textit{Listen} action in the \textit{Tiger} problem. Although we discussed the case in the latter problem as the agent being trapped by its belief, the underlying reason in this case is yet obvious. The direct reason is that the \textit{Check} action has its value increased in belief nodes of the decision tree that makes itself a little more favorable in Monte-Carlo tree search process. Behind the increased value of the action, a possible explanation is that more \textit{Check}-ings are needed to be done in order to raise the agent's certainty about a rock's state, which is potentially attractive to the agent as the \textit{Good} rock state brings back a reward. 

Apart from above analysis on direct statistics, we also try to investigate the causation between the deceptive kernel and agent's decisions. We hope to find connections between each false observation generated by the deceptive kernel to a potential mistake made by the agent. Now we briefly describe a probable route that leads from deception to mistake. The agent primarily receives a false observation which directly or indirectly changes its originally correct belief (regarding a rock) to be incorrect. Later the agent manages to reach that rock position but makes the wrong decision by either sampling a bad rock or ignoring a good rock. We can notice three key stages in this route. First, a belief-change is (potentially) triggered by the deceived observation; second, the agent later visits the rock position; lastly, the agent makes a mistake at the position. In Table \ref{tab3} we collect the statistics regarding the number of observations, the number of belief-changes, and further the number of changes that can be attributed to the false observation and deceived observation. 

\begin{table}[]
\centering
\caption{Statistics of Observations and Belief Changes}\label{tab3}
\begin{tabular}{c|cccc}
\toprule
(Number of)                  & \head{Baseline} & \head{Prob} & \head{Rand} & \head{Oppo} \\
\midrule
Observations                 & 2186            & 2252        & 2872        & 2784 \\ 
Belief changes               & 561             & 661         & 1197        & 639  \\ 
Belief change - false obs    & 28              & 129         & 546         & 639  \\ 
Belief change - deceived obs & 0               & 108         & 517         & 639  \\
\bottomrule
\end{tabular}
\end{table}

Table \ref{tab3} clearly presents that belief changes induced by our artificial deception take up a significant part in experiments with \textit{Prob} and \textit{Rand} kernel, while in the meantime the ones induced by normal observations remain at its baseline amount. With \textit{Oppo} kernel, not surprisingly all the observations and belief changes are deceived. 

From this result, we further investigate all of agent's observations correlated with agent's later action regarding the targeted rock and give the investigated statistics in Table \ref{tab4}. In Table \ref{tab4}, we categorize each observation in two major classes $\{Normal, Deceived\}$ and one of five subclasses $\{TP, TN, FP, FN, Ignore\} $. Assume an observation received by the agent is $o$ and is targeting rock $i$ whose position is $p_i$. Then we define $o$ is \textit{T} if the observation itself is positive (regardless of being deceived or not) and \textit{F} otherwise. We also define $o$ is \textit{P} if the agent samples the rock $i$ after observing $o$; define $o$ is \textit{N} if the agent after observing $o$ visits $p_i$ but does not sample it. Particularly, we also define $o$ is of case \textit{Ignore} if $o$ is positive but the agent after observing it does not visit $p_i$. 

Then for observations being \textit{Deceived} and of case \textit{TP} or \textit{FN}, we assume there is a causation between deceptive observation and agent's mistake as it either samples a bad rock (\textit{TP}) or visit but ignore a good rock (\textit{FN}). For a deceptive observation of case \textit{TN} or \textit{FP}, we say the deception fails as the agent ignores the deception and manages to either sample the good rock or avoid the bad. On the other hand, for a \textit{Normal} (not deceived) observation of case \textit{TP} or \textit{FN}, it implies the causality that observation leads to agent's right choice. Otherwise, a normal observation being \textit{TN} or \textit{FP} implies the agent fails to decide correctly based on the observation or is affected by environment error. 

\begin{table}[]
\centering
\caption{Distribution of Categorized Observations}\label{tab4}
\begin{tabular}{c|l|cccc}
\toprule
\head{Observation}        & \head{Subclass}         & \head{Baseline}     & \head{Prob}   & \head{Rand}   & \head{Oppo} \\
\midrule
\multirow{7}{*}{Normal}   & TP                      & 556                 & 510           & 281           & 0    \\  
                          & FN                      & 1169                & 1087          & 836           & 0    \\  
                          & TN                      & 25                  & 31            & 25            & 0    \\  
                          & FP                      & 10                  & 6             & 5             & 0    \\  
                          & Ignore                  & 426                 & 360           & 345           & 0    \\ \cline{2-6} 
                          & Total                   & 2186                & 1994          & 1492          & 0    \\ \cline{2-6} 
                          & (TP+FN)/(TP+FN+TN+FP)   & 0.98                & 0.98          & 0.97          & N/A  \\
\midrule
\multirow{7}{*}{Deceived} & TP                      & N/A                 & 0             & 8             & 0    \\  
                          & FN                      & N/A                 & 85            & 513           & 1134 \\ 
                          & TN                      & N/A                 & 2             & 5             & 29   \\  
                          & FP                      & N/A                 & 81            & 158           & 0    \\ 
                          & Ignore                  & N/A                 & 90            & 696           & 1621 \\ \cline{2-6} 
                          & Total                   & N/A                 & 258           & 1380          & 2784 \\ \cline{2-6} 
                          & (TP+FN)/(TP+FN+TN+FP)   & N/A                 & 0.51          & 0.76          & 0.98 \\
\bottomrule
\end{tabular}
\end{table}

We see in Table \ref{tab1} that the ratio $(TP+FN)/ (TP+FN+TN+FP) $, representing the `efficiency' of deceived observations, is apparently not proportional to the one of normal observations. If the agent receives a deceived observation by \textit{Prob} kernel and it later visits the rock position, there is around 50\% chance that the agent will make wrong decision with the rock. If the kernel switches to \textit{Rand} then the chance increases to 76\%. This result backs up our assumption that observations generated by deceptive kernel contribute to, if not lead to, significantly more mistakes by the agent. 

\begin{figure}[h!]
\centering
\includegraphics[width=5.0in]{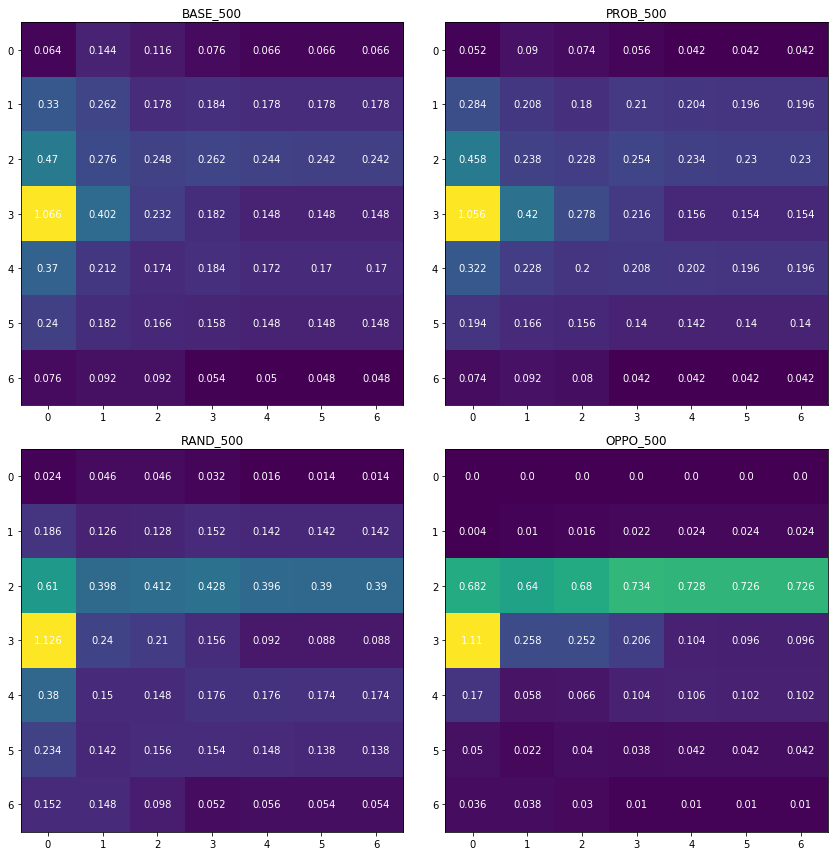}
\caption{Heatmaps of the agent's occurrence} \label{fig5}
\end{figure}

From another perspective, we plot the heatmap showing agent's average occurrence at every position in the map as Figure \ref{fig5}. It is shown that in baseline model the agent is the most explorative by having a relatively more dispersive occurrence over the entire map. As the deceptive kernel gets more effective, the agent tends to be more conservative regarding its path choice. With \textit{Oppo} kernel feeding entirely wrong observations, the agent almost fixes a fastest path to exit from the map as soon as possible.

\subsubsection{Further Experiment with Costly Deception in RockSample} \hfill\\

Mentioned in section \textbf{Cost for Deception}, we also test our thought on introducing a little reward $r_d = 1$ to the agent for every time it receives a deceptive observation. In Table 5, we provide statistics of the relative experiments as parallel contents to Table 2. The undiscounted returns under deceptive kernel are not surprisingly higher than their parallel results in original setting, since the agent is rewarded by being deceived. However, an interesting result is that, regarding the discounted return under \textit{Rand} kernel, a much higher average number (36.14) of \textit{Check}-ings taken does not produce a proportionate return (13.85) as compared with its parallel result (5.47; 8.63). The introduction of this tiny reward to deception is seemingly a beneficial thing to the agent, and consequently it induces agent's greater favor and even attachment to it. However, excessive investment on the certain action does not return at the same level as the discount factor diminishes most of the agent's effort. 

\begin{table}[h!]
\centering
\caption{Parallel results of experiments on costly deception}\label{tab5}
\begin{tabular}{c|c|c|c|c|c}
\toprule
                      & {\head{Baseline}} & {\head{Prob}} & {\head{Prob rewarded}} & {\head{Rand}} & \head{Rand rewarded} \\
\midrule
Undiscounted return   & 23.70            & 20.80             & 24.69            & 15.30              & 41.33      \\ 
Discounted return     & 15.21            & 13.21             & 14.15            & 8.63               & 13.85      \\ 
Average sampled rocks & 1.37             & 1.08              & 1.32             & 0.53               & 1.40       \\ 
Average checkings     & 4.61             & 4.46              & 8.36             & 5.47               & 36.14      \\ 
Average steps         & 14.57            & 13.85             & 18.31            & 14.05              & 46.32      \\ 
\bottomrule
\end{tabular}
\end{table}

\section{Conclusion and Future Work}
In this paper, a deceptive approach based on observation model in POMDP problems has been introduced to study its potential influence on agent's behavior and return. We have provided three general implementations as deceptive kernel functions that exert deception on agent's observation. In section 3, theoretical reasoning has been given about the kernel's effect on primarily agent's belief state and potentially the decision-making process, with certain limitations in POMDP model's settings and adopted algorithms. Experimental results in section 4 supports our assumption and analysis about the kernel's detrimental effect on the return and causalities between deceptive observations and the increase of agent's mistakes. We also test a new setting of kernel in section 4.2 by introducing rewarding deceptions and reach an interesting result for the \textit{Rand} kernel. Further work can be invested on introducing a general framework of deceptive kernel function applicable to a generic POMDP problem, improving the current kernel by raising its efficiency and targeting ability while lowering its cost, and exploring more of the deception's potential in adversarial games.  
%
%
%
%
%
%

%
%

%
%
%
\bibliographystyle{splncs04}
\bibliography{Deceptive_Kernel_Function_on_Observations_of_Discrete_POMDP}

\end{document}